%% file: main.tex
\def\year{2020}\relax
\documentclass[letterpaper]{article} 
\usepackage{aaai20}  
\usepackage{times}  
\usepackage{helvet} 
\usepackage{courier}  
\usepackage[hyphens]{url}  
\usepackage{graphicx} 
\usepackage{graphicx}  
\usepackage{amsmath}
\usepackage{latexsym}
\usepackage{multirow}
\usepackage{booktabs}
\newcommand{\citet}[1]{\citeauthor{#1} \shortcite{#1}}
\newcommand{\citep}{\cite}

\DeclareTextSymbolDefault{\textquotedbl}{T1}
\providecommand{\tabularnewline}{\\}

\title{Towards Lingua Franca Named Entity Recognition with BERT}
\author{
  Taesun Moon\thanks{Both authors contributed equally.} \and Parul Awasthy\footnotemark[1] \and Jian Ni \and Radu Florian \\
    IBM Research AI\\
    Yorktown Heights, NY 10598\\
    {\{tsmoon, awasthyp, nij, raduf\}@us.ibm.com}\\
} 
\begin{document}

\maketitle

\begin{abstract}
Information extraction is an important task in NLP, enabling the
automatic extraction of data for relational database
filling. Historically, research and data was produced for English text,
followed in subsequent years by datasets in Arabic, Chinese
(ACE/OntoNotes), Dutch, Spanish, German (CoNLL evaluations), and many
others. The natural tendency has been to treat each language as a different
dataset and build optimized models for each. In this paper we
investigate a single Named Entity Recognition model, based on a
multilingual BERT, that is trained jointly on many languages
simultaneously, and is able to decode these languages with 
better accuracy than models trained only on one language. To improve the initial model, we study the use of regularization strategies such as multitask learning and partial gradient updates. In addition to being a single model that can tackle multiple languages (including code switch),
the model could be used to make zero-shot predictions on a new
language, even ones for which training data is not available, out of the box. The results show that this model not only performs competitively
with monolingual models, but it also achieves state-of-the-art
results on the CoNLL02 Dutch and Spanish datasets, OntoNotes Arabic
and Chinese datasets. Moreover, it performs reasonably well
on unseen languages, achieving state-of-the-art for
zero-shot on three CoNLL languages.
\end{abstract}

\input{01-intro.tex}
\input{02-prior.tex}

\input{tables/table_multitrain.tex}
\input{03-task.tex}
\input{tables/ontonotes.tex}
\input{tables/table_zeroshot.tex}

\input{04-exps.tex}
\input{05-dis.tex}
\input{06-exps-frz.tex}
\input{07-exps-mult.tex}

\input{08-conc.tex}

\bibliographystyle{aaai}
\bibliography{moon,jian,sil,emnlp-ijcnlp-2019}
\end{document}

%% file: 01-intro.tex
\section{Introduction} \label{sec:Introduction} 

Named entity recognition (NER) is an important
task in information extraction and natural language processing. Its
main goal is to identify, in unstructured text, contiguous typed references
to real-world entities, such as persons, organizations, facilities,
and locations. It is very useful as a precursor to identifying semantic
relations between entities (to fill relational databases), and events
(where the entities are the events' arguments). 

The vast majority of NER research focuses on one language at a time, building and tuning different models for each language with monolingual data. The models have evolved from rule-based to statistical models, closely following
general trends in NLP: from Winnow and Transformation-based learning, to
Maximum Entropy, SVMs, and CRF-based models, voted-perceptron, to
deep neural networks, and more recently using pre-trained models on large amounts of unlabeled data (ELMo \cite{ELMo18}, BERT \cite{BERT18}). However, the single-language aspect of research stayed constant, with few
exceptions described later in Prior Work. Monolingual models
will not be able to share resources across languages, requiring large 
amounts of manually labeled data in each language.

In this paper, we posit the hypothesis that it is not only possible,
but certainly beneficial, to share data across different languages
with the proper statistical model, even in cases where the languages
are not from the same language family or using the same script. Building upon a BERT multilingual model \cite{BERT18}, we show that (1) a model can be
built to handle multiple languages at the same time, (2) the joint
model performs better than a model built on the same architecture but
with only one language, and (3) the model can be used to perform
\emph{0-shot (language-wise) NER}\footnote{Basically, a mode
  where the NER model performs inference on a language that it was not
  trained on.} with reasonable accuracy. Such a cross-language model
has many advantages in a production environment: simplified deployment
and maintenance, easy scalability, the same memory/CPU/GPU footprint,
and, of course, the 0-shot ability.
However there are limitations to the zero-shot learning framework from
the perspective of the standard train/evaluation/model selection process
that 4) we attempt to overcome with several modifications which push the
SOTA even further. In particular, we examine extensions to the training framework, including partial gradient updates and incorporating additional tasks such as cloze prediction and language prediction.



%% file: 02-prior.tex
\section{Prior Work}\label{sec:Prior-Work}
Named entity recognition and its successor, mention detection, have
a vast history in NLP - a full description is beyond the scope of
this paper. We will touch on the deep learning research that is directly related to the results presented here.

\citet{collobert08} was the first modern approach to sequence
classification, including NER, that used a convolutional neural network architecture, advancing the state-of-the-art (SotA) in English
CoNLL.  \citet{LampleBSKD16} introduced -- what has become the
standard baseline -- Bidirectional LSTM (Bi-LSTM) networks to advance the SotA
NER performance on the CoNLL datasets, building 4 models, one for each
language.

2018 saw the introduction of strong language-model pretrained models,
first with ELMo\cite{ELMo18}, then with BERT\cite{BERT18}. \citeauthor{ELMo18} used large amounts of unlabeled English data to train a 3-layer LSTM
network with inputs from character embeddings, that are then used as
input to standard Bi-LSTM networks to obtain the state-of-the-art
results in many tasks. In a similar fashion, \citet{BERT18} trained a
transformer-based architecture on large amounts of unlabeled text,
with a cloze and next sentence prediction objectives, then feeding the
sentence/paragraph embeddings to a linear feedforward layer, again
surpassing the SotA in many tasks.

\citet{akbik-etal-2018-contextual} extends the ELMo framework by
computing Bi-LSTM sequences at character level for the entire
sentence, then combines the token aligned pieces to feed into a
bidirectional LSTM layer, together with the word embeddings, and
obtaining SotA results on CoNLL and OntoNotes. 
The current SotA in English CoNLL'03 is obtained by
\cite{Baevski19-cloze5}, who pretrain a two-tower attention model on a
cloze task and then apply the resulting embedding to obtain SotA
results on the GLUE tasks, and obtain 93.5 $F_1$ on the English CoNLL'03
dataset.
\begin{figure}
\begin{centering}
\vspace*{-4mm}
\hspace*{-4mm}
\includegraphics[scale=0.37]{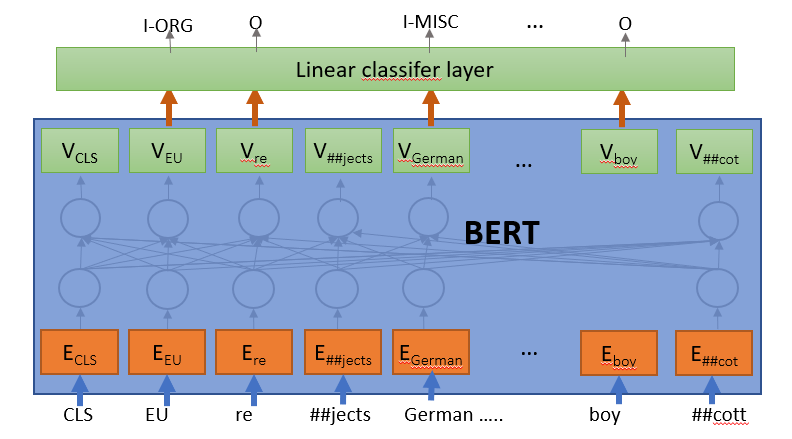}
\par
\end{centering}
\caption{BERT-ML Architecture\label{figure:BERT}}
\vspace*{-5mm}
\end{figure}

\subsubsection*{Multilingual Work}
There are a few recent investigations into building models that can
handle multiple languages at the same time. \citet{sil2015-tac} trained
a joint mention detection model on English and Spanish, resulting in
better performance on the Spanish
data. \citet{akbik-etal-2018-contextual} did experiments by training
Flair on all CoNLL'02 and '03 languages and providing one model on
their github page of their system, Flair \citet{akbik19:flair} -- this
system is the closest to the model described here, and we compare our
models with theirs. The main difference is that our system can be
easily run on a new language that doesn't share the input script, as
we will show in the experiment section.

In related work, \citet{Xie2018} aligned monolingual embeddings from
English to Spanish, German, and Dutch, and then translated the English
CoNLL dataset into these languages, and built a self-attentive
Bi-LSTM-CRF model using the translated languages, creating 0-shot NER
systems.  \citet{pires-etal-2019-multilingual} used multilingual BERT and techniques similar to our zero-shot baseline to obtain SotA numbers for zero-shot on all four CoNLL languages. 

\subsubsection*{Multi-task Learning}
Multi-task learning (MTL) has been used successfully across many applications of machine learning for many years now. \citet{Caruana1997MultitaskL} gives a nice overview of MTL, stating: ``MTL improves generalization by leveraging the domain-specific information contained 
in the training signals of related tasks." MTL has also been popular in natural language processing. \citet{collobert08} used MTL effectively to jointly learn various NLP tasks like part-of-speech tagging, chunking, named entity recognition, semantic role labeling, etc. by sharing a single network for the tasks. Since then there have been works implementing MTL as a hard shared network, or soft parameter sharing \cite{Duong2015LowRD} where each task has its own model with its own parameters.
The distance between the parameters of the model is then regularized in order to encourage the
parameters to be similar. The BERT models are also trained using MTL, where they share one network to learn two tasks simultaneously, one a Cloze style language model and the other a model for next sentence prediction.

%% file: tables/table_multitrain.tex
\makeatletter
\setlength{\@fptop}{0pt}
\makeatother

\begin{table}[ht!]
\setlength\tabcolsep{2pt} 
\begin{tabular}{|p{4cm}|c|c|c|c|}
\hline 
{\small{}{System}} & {\small{}{English}} & {\small{}{Spanish}} & {\small{}{German}} & {\small{}{Dutch}}\tabularnewline
\hline 
{\small{}CoNLL'02 and '03} & {\small{}{88.6}} & {\small{}{81.4}} & {\small{}{72.4}} & {\small{}{77.1}}\tabularnewline
\hline 
{\small{}\citet{Baevski19-cloze5}} & {\small{}\textbf{93.5}} & {\small{}{-}} & {\small{}{-}} & {\small{}{-}}\tabularnewline
\hline 
{\small{}\citet{akbik-etal-2018-contextual}} & {\small{}{93.2}} & {\small{}{-}} & {\small{}\textbf{88.2}} & {\small{}{90.4}}\tabularnewline
\hline 
{\small{}\citet{BERT18}} & {\small{}{92.6}} & {\small{}{-}} & {\small{}{-}} & {\small{}{-}}\tabularnewline
\hline 
{\small{}\citet{LampleBSKD16}} & {\small{}{90.9}} & {\small{}{85.8}} & {\small{}{78.8}} & {\small{}{81.7}}\tabularnewline
\hline 
{\small{}Flair-ML} & {\small{}{92.2}} & {\small{}{86.6}} & {\small{}{74.9}} & {\small{}{88.9}}\tabularnewline
\hline \hline
{\small{}{BERT-SL (this work)}} & {\small{}{91.2}} & {\small{}87.5} & {\small{}82.7} & {\small{}{90.6}}\tabularnewline
\hline 
{\small{}{BERT-ML (this work)}} & {\small{}{91.3}} & {\small{}\textbf{87.9}} & {\small{}{83.3}} & {\small{}\textbf{91.1}} \tabularnewline
\hline 
\end{tabular}{\small{}{\small{}\vspace*{2mm}}}%

{\small{}\caption{Single and multi language F$_{1}$ on CoNLL'02, CoNLL'03. Flair-ML is the system described in \cite{akbik-etal-2018-contextual}, trained multilingually, available from \cite{akbik19:flair}\label{table:Multitrain}.}
}{\small\par}
\end{table}

%% file: 03-task.tex
\section{Task and Framework}
Most neural-based NER systems start building upon word embeddings that capture language use. This is usually
achieved by pretraining word embeddings with various techniques, such as
CBoW, skip-gram, ELMo, BERT, RoBERTa, XLNet, etc. The challenge of building on multiple languages, especially if they have different scripts, is to be able to
train these word embeddings in such a way that similar words in different
languages have similar representations\footnote{Note that the
  similarity here can either be standard synonymy or just task
  similarity. From an NER perspective "John" and "Mary" are
  similar.}. \citet{ni-etal-2017-weakly} used representation projection
between pairs of languages (e.g. Spanish to English) to be able to use
the English NER tagger on Spanish data, with moderate success, but the
models themselves were not benefiting from data in both (or many)
languages.

We hypothesize here that the BERT word piece embeddings are well
aligned by the pretraining, so that a BERT-based system will be able to
successfully train on multiple languages, and will also be able to
perform zero-shot inference on text from a new language. Furthermore,
this approach has the great property that it will have exactly the
same memory/CPU/GPU footprint, regardless of the number of languages
it can do inference on.

This work is based on \cite{BERT18} BERT framework, using the
pretrained \textit{bert-base-multilingual-cased} embeddings for all
models. These embeddings are trained on the top 100 languages with largest
Wikipedia \cite{google-research18:_tensor_bert}. As the languages'
sizes vary, the data is sampled by using exponentially smoothed
weighing to balance the language representations in the training
corpus. These embeddings comprise a shared WordPiece vocabulary
of 110k pieces spanning many non-European scripts including Arabic and Chinese.

We approach NER in a standard fashion: a sequence labelling task which
assigns a tag to each word based on its context. Given a sentence
$\left\{w_{1},w_{2},....w_{n}\right\}$, we feed it to the BERT model
to obtain contextual BERT embeddings for each word as
\{$v_{1},v_{2},...v_{n}$\}, capturing each word's context via many
attention heads in each of its layer. These embeddings are then fed to
a linear feed forward layer to obtain labels
\{$y_{1},y_{2},...y_{n}$\} corresponding to each each word piece (see Figure
\ref{figure:BERT}). This entire network is trained with each epoch
thereby fine-tuning the BERT embeddings for the NER task. We are using
an IOB1 encoding of the entities \cite{sang99representing}, as it
performed best in preliminary results.

We use the HuggingFace PyTorch implementation of BERT
\cite{huggingface-github19} and the BERT WordPiece Tokenizer. We
follow the recipe in \cite{BERT18} for building named entity taggers:
to convert the NER tags from tokens to word pieces, we assign the tag
of the token to its first piece, then assign the special tag 'X' to
all other pieces. No prediction is made for \textquotedbl
X\textquotedbl{} tokens during training and testing. Figure
\ref{figure:BERT} shows both the architecture of the proposed model,
and the NER annotation style.

%% file: tables/ontonotes.tex
\makeatletter
\setlength{\@fptop}{0pt}
\makeatother

\begin{table}[ht!]
\begin{center}
\begin{tabular}{|l|c|c|c|}
\hline 
{\small{}{System}} & {\small{}{English}} & {\small{}{Arabic}} & {\small{}{Chinese}}\tabularnewline
\hline 
{\small{}{\citeauthor{akbik-etal-2018-contextual}}} & {\small{}\textbf{89.7}} & {\small{}{-}} & {\small{}{-}}\tabularnewline
\hline 
{\footnotesize{}{\citeauthor{clark-etal-2018-semi}}} & {\small{}{88.8}} & {\small{}{-}} & {\small{}{-}}\tabularnewline
\hline 
{\small{}{\citet{ghaddar-langlais-2018-robust}}} & {\small{}{88.0}} & {\small{}{-}} & -\tabularnewline
\hline 
{\small{}{\citet{pradhan-etal-2013-towards}}} & {\small{}82.4} & {\small{}{68.0}} & {\small{}{71.8}}\tabularnewline
\hline \hline
{\small{}{BERT-SL (this work)}} & {\small{}{87.9}} & {\small{}{68.7}} & {\small{}{72.9}}\tabularnewline
\hline 
{\small{}{BERT-ML (this work)}} & {\small{}{88.3}} & {\small{}\textbf{69.9}} & {\small{}\textbf{74.1}}\tabularnewline
\hline 
{\small{}{BERT-SL$^{Eng}$ 0-shot}} & {\small{}{87.9}}&{\small{}{10.7}} &{\small{}65.2} \tabularnewline
\hline
{\small{}{BERT-ML$^2$ 0-shot}} & {\small{}61.0} &  {\small{}12.6}& {\small{}65.6} \tabularnewline
\hline 
{\small{}{WordPiece fertility}} & {\small{}{1.8}}  & \small{}{7.42}& {\small{}{2.31}}\tabularnewline
\hline 
\end{tabular}{\small\par}

{\small{}\caption{Single language, multi-language and 0-shot F$_{1}$ on OntoNotes. WordPiece fertility is the average number of BERT wordpieces per token in a given language. \label{table:Ontonotes}}
}{\small\par}
\end{center}
\end{table}

%% file: tables/table_zeroshot.tex
\begin{table}[ht!]
{\small{}}%
\setlength\tabcolsep{2pt} 
\begin{tabular}{|p{4cm}|c|c|c|c|}
\hline 
{\small{}{System}} & {\small{}{{German}}} & {\small{}{English}} & {\small{}{{Spanish}}}  & {\small{}{{Dutch}}}\tabularnewline
\hline 
{\small{}{\citet{Xie2018}}} & {\small{}{}56.9} & {\small{}-} & {\small{}{}72.4}  & {\small{}{}71.3}\tabularnewline
\hline 
{\small{}{\citet{pires-etal-2019-multilingual}}} & {\small{}{}69.74} & {\small{}90.70$^*$} & {\small{}{}73.59}  & {\small{}{}77.36}\tabularnewline
\hline 
{\small{}{BERT-SL$^{Eng}$ 0-shot}} & {\small{}{69.42}} & {\small{}-} & {\small{}{73.62}}  & {\small{}{78.61}}\tabularnewline
\hline 
{\small{}{BERT-ML$^3$ 0-shot}} & {\small{}\textbf{70.23}} & {\small{}{}72.57} & {\small{{}\textbf{77.17}}}  & {\small{}{79.76}}\tabularnewline
\hline 
{\small{}{CL}} & {\small{}{}67.04} & {\small{{}{73.30}}} & {\small{75.29}} & {\small{}\textbf{81.76}}\tabularnewline
\hline 
{\small{}{CL+LI}} & {\small{}{}68.94} & {\small{\textbf{74.28}}} & {\small{73.68}} & {\small{}{80.78}}\tabularnewline
\hline 
{\small{}{LI}} & {\small{}{}71.28} & {\small{{}{73.87}}} & {\small{75.40}} & {\small{}{80.06}}\tabularnewline
\hline 
{\small{}{PC}} & {\small{{}{67.91}}} & {\small{73.72}} & {\small{}{}73.68} & {\small{}{80.28}}\tabularnewline
\hline 
{\small{}{PC+LI}} & {\small{}{}66.31} & {\small{{}{74.79}}} & {\small{74.12}} & {\small{}{79.53}}\tabularnewline
\hline 
\end{tabular}{\small{} }%

{\small{}\caption{Zero-shot F$_{1}$ scores for models trained on
CoNLL languages.  BERT-ML$^k$ is the BERT-ML system trained on $k$ languages. Numbers from \cite{pires-etal-2019-multilingual} are the from the English row in their Table 1. For details on the CL, CL+LI,LI,PC,PC+LI systems, see Section~\ref{sec:exp:mult}\label{table:zeroshot}}; English numbers are not 0-shot in this setup.
}{\small\par}
\end{table}

%% file: 04-exps.tex
\section{Experiments: Baselines} \label{sec:exp:zero-shot}

In this section, we examine our approach in the base zero-shot setting, where we use the cased multilingual BERT embeddings out of the box and only fit and fine tune it on the set of languages which are not the target language, and then evaluate on the target language.

\subsection{Data and Experimental Setup}

We train/evaluate our models on two datasets, CoNLL NER '02/'03 \cite{TjongKimSang:2002:ICS:1118853.1118877} and
OntoNotes 5.0 \cite{ontonotes5}.  The CoNLL datasets consist of
newswire data in four European languages Spanish, Dutch, English and
German
and is annotated with four entity types (PER, LOC, ORG, MISC).

OntoNotes 5.0 is a more challenging corpus, containing three languages
that do not share scripts: Arabic, Chinese and English and contains 18
NER types.

For most experiments we use learning rate of $5^{-5}$ and batch size of
32 or 100 and train for 20 epochs
. We tune the hyper-parameters on the development and choose the model
with best F1 score on the development set. We report the test numbers
on that model (the training data does not include the development
test). All test numbers shown are an average of three runs. Models
were trained on a desktop PC with two Nvidia 2080Ti cards; training
times ranged from 0.5 to 1.5 hours.

\subsection{Comparison Methods}

\textbf{\textit{Monolingual Training}}

As a baseline, we train models
on all four CoNLL languages and three OntoNotes languages separately
\footnote{Only the \textit{base} model was released for multilingual
  embeddings; it is not unreasonable to expect a \textit{large} model
  to yield better results.}.

\vspace*{1mm}\noindent \textbf{\textit{Multilingual Training}}

We train a joint multilingual model on CoNLL data
and one with the OntoNotes data. To train multilingually, we randomly
mix sentences from all languages and train on them in each epoch.
Both are monolingual and multilingual models are based on the same
architecture, the only difference being the data used to train the
model.


\vspace*{1mm}\noindent\textbf{\textit{Zero-shot Inference}}

To test the power of the trained model to perform zero-shot inference,
we train models, in a round-robin fashion, on three of the CoNLL
languages and test on the fourth, and similarly on two OntoNotes
languages and test on third. 

\subsection{Experimental Results and Analysis}

The results for the monolingual and multilingual experiments are shown
in Table \ref{table:Multitrain} for CoNLL and Table \ref{table:Ontonotes} for Ontonotes. For both CoNLL and OntoNotes, the
model that was trained on all the languages (BERT-ML) performs better
than the model trained on one language at a time (BERT-SL), in all
cases, showing that the BERT-ML model is able to benefit from
information coming from all the languages to improve each individual
language, validating our first two hypotheses: that we can
effectively build a multi-language model, and that it performs better
than individual models. Relative to other published results, the
Bert-ML CoNLL model obtains SotA results for Spanish and Dutch and is
ranked second for German, and the Bert-ML OntoNotes model obtains SotA
for Chinese and Arabic and is ranked third for English. Also, note
that the BERT-ML model performs better than the Flair-ML model, except
for English, in spite of the fact that the size/architecture of the
former is the same as for every language \footnote{The Flair-ML model
  grows linearly with each added language, due to needing the word
  embeddings from that language.}\footnote{We realize that the English
  performance is lower than reported in \cite{BERT18}. Despite our
  best efforts, we were not able to replicate that performance, or
  even find other reported success in the community in replicating the
  92.4/92.8 numbers, though many other unsuccessful attempts were
  reported, which leads us to believe that parts of the system were
  not fully described.}.

To validate the third hypothesis, Table \ref{table:zeroshot} shows the
results for the zero-shot experiments. The BERT-ML system obtains good
results on Spanish, German, and Dutch (CoNLL) and Chinese (OntoNotes),
without seeing any training data in that language. For the CoNLL
languages, the performance is very close with the results of the top
systems participating in the original competitions, on Dutch being 2F
higher than the winning system. Also shown in the Table are the
(previous) SotA result in 0-shot cross language NER on CoNLL. The
BERT-SL model trained on English alone has good performance
across the board.

For OntoNotes, Table \ref{table:Ontonotes}, among all the languages, Arabic does not perform great in a 0-shot
scenario - one only gets 12.6F. Looking at the wordpiece dictionary in
the BERT tokenizer, we found out that Arabic is not well
represented. As we show in the last row, one Arabic
token gets split into 7.42 word pieces on average, while all other
languages are in a range of 1.8-2.5. This has a damaging effect due to
lengthening of the tokens, resulting in significantly worse behavior in
Arabic than in the other languages. Despite this, the model is
still able to incorporate information from English and Chinese, when
trained together, and improve 1.2F over the Arabic only model.


%% file: 05-dis.tex
\section{Discussion: Zero-shot\label{sec:exp:zer}}

\begin{figure}
\begin{center}
\includegraphics[scale=0.5]{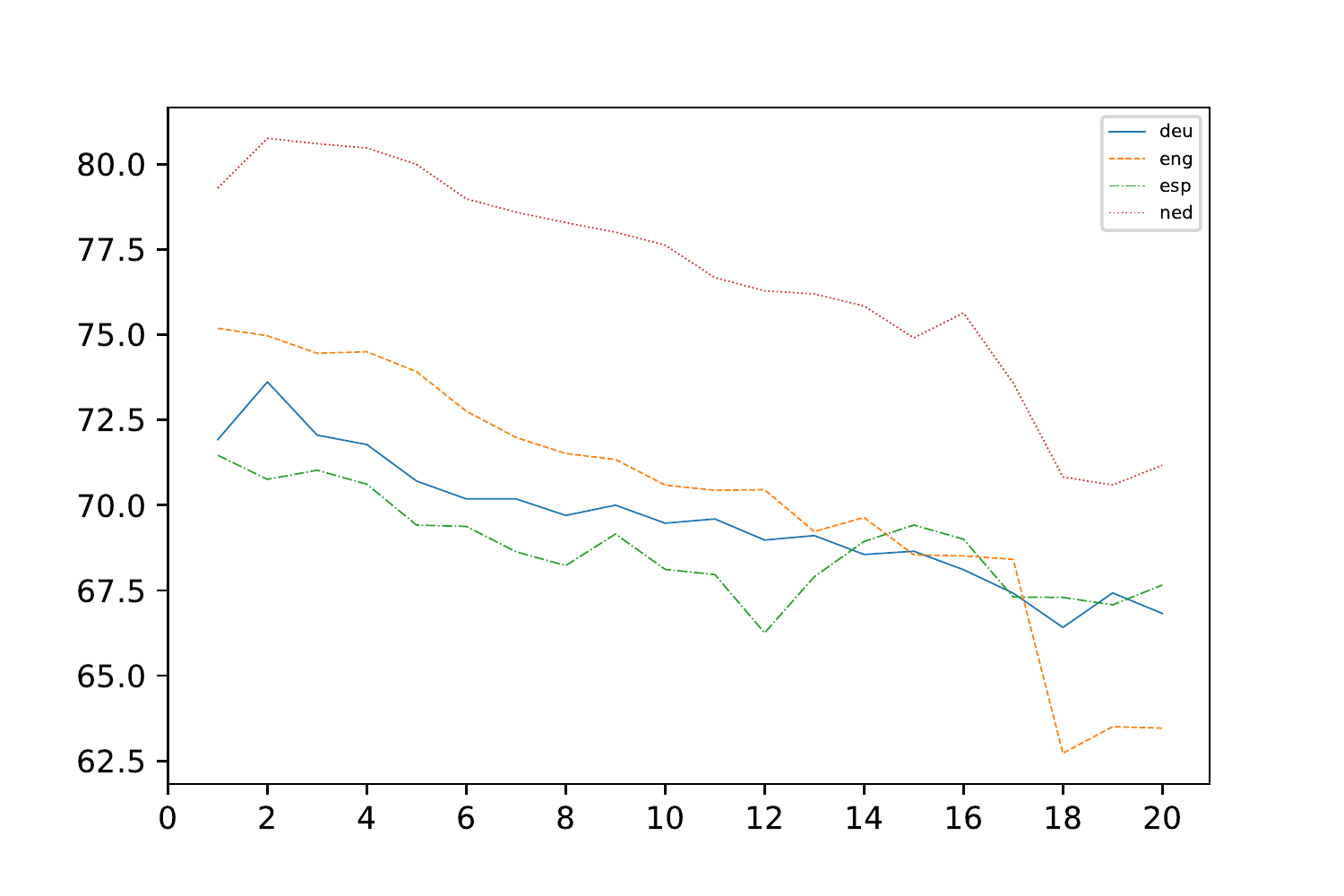}
\end{center}
\caption{Performance of zero-shot BERT multilingual over epochs on CoNLL dev by
  target language. Average of 5 runs}
\label{fig:conll_fscore_epochs_by_lang_csd}
\end{figure}

The results in table \ref{table:zeroshot} for BERT-ML 0-shot on the
CoNLL data are impressive, especially given the fact that no annotated
data in the respective target languages are used. However, when we
examine the performance of these models on the respective development
data sets over the training epochs, a peculiar picture emerges (Fig.
\ref{fig:conll_fscore_epochs_by_lang_csd}). Considering the case where the target language
is English while training the model using the Dutch, German and
Spanish labeled data, we evaluate on the CoNLL English development set after
every epoch. Averaging the overall F-score over five runs, we
clearly see a downward trend of the F-score as training progresses. This
is outside the norm for most statistical learning problems with labeled
data, where we expect to see gradual increases in the F-score as
learning progresses and then possibly a plateau and not the continued downward
trajectory seen in this graph.

\input{tables/conll_frz_cased_dev.tex}

\input{tables/conll_frz_cased_tst2.tex}

One reason for why this might be happening stems from our initial hypothesis, that the multilingual BERT embeddings are well-aligned. But it might be that this is true only at the beginning before we begin finetuning the BERT model. While finetuning on the labeled data, overtraining relative to the target language occurs much sooner than in standard settings. This makes sense because we are actually not doing any training with the target language and thus the learning objective for the languages we (hypothetically) have labels for is dissociated from the goal we have in mind, which is to do well in NER on the target language.

To overcome this pathology in the F-score trajectory over epochs,
we propose two training enhancements. The first is to limit learning to a subset of the BERT transformer architecture, i.e. to freeze 1 or more of the lower layers of the transformer during learning such that no back-propagation and updates occur on these lower layers. The second is to incorporate the target language in an unsupervised manner to enable transfer learning for the target language while also simultaneously optimizing the model for the languages where we do have labeled data.

%% file: tables/conll_frz_cased_dev.tex
\begin{table}
\begin{center}
\begin{tabular}{lrrrr}
\toprule
\#frz &   German & English & Spanish & Dutch \\
\midrule
-1  & 77.28 & 77.36 & 75.87 & 82.41 \\ 
 5  & 79.06 & 77.78 & 74.42 & 83.58 \\ 
 7  & 77.92 & 79.38 & 73.48 & 83.21 \\ 
 11 & 67.87 & 72.41 & 69.58 & 78.40 \\ 
 12 & 53.84 & 63.43 & 58.86 & 66.14 \\ 
\bottomrule
\end{tabular}
\caption{Cased multilingual zero shot on CONLL development data. \#frz
  corresponds to number of frozen layers during training. -1 is where
  nothing is frozen and is the standard setting. 0 is where only the
  embedding is frozen. 1 is embedding and first layer is frozen and so
  forth. 12 is where all bert layers are frozen and only the output
  layer is tuned.}
\label{tab:conll_frz_cased_dev}
\end{center}
\end{table}

%% file: tables/conll_frz_cased_tst2.tex
\begin{table}
\begin{center}
\begin{tabular}{lrrrr}
\toprule
\#frz     & German & English & Spanish & Dutch \\
\midrule
-1  & 70.34 & 72.57 & 77.17 & 79.76 \\ 
3  & \textbf{72.44} & 73.61 & 76.53 & \textbf{83.35} \\
 7  & 71.28 & \textbf{76.20} & 75.50 & 81.01 \\ 
 11 & 59.20 & 69.20 & 70.72 & 75.05 \\ 
 12 & 43.45 & 59.28 & 58.43 & 64.17 \\ 
\bottomrule
\end{tabular}
\caption{Zero-shot BERT multilingual on CONLL TEST data. \#frz
  corresponds to number of frozen layers during training. -1 is where
  nothing is frozen and is the standard setting. 0 is where only the
  embedding is frozen. 1 is embedding and first layer is frozen and so
  forth. 12 is where all bert layers are frozen and only the output
  layer is tuned.}
\label{tab:conll_frz_cased_tst2}
\end{center}
\end{table}

%% file: 06-exps-frz.tex
\section{Experiments: partial updates}
\label{sec:exp:part-up}


\begin{figure}
\includegraphics[scale=0.5]{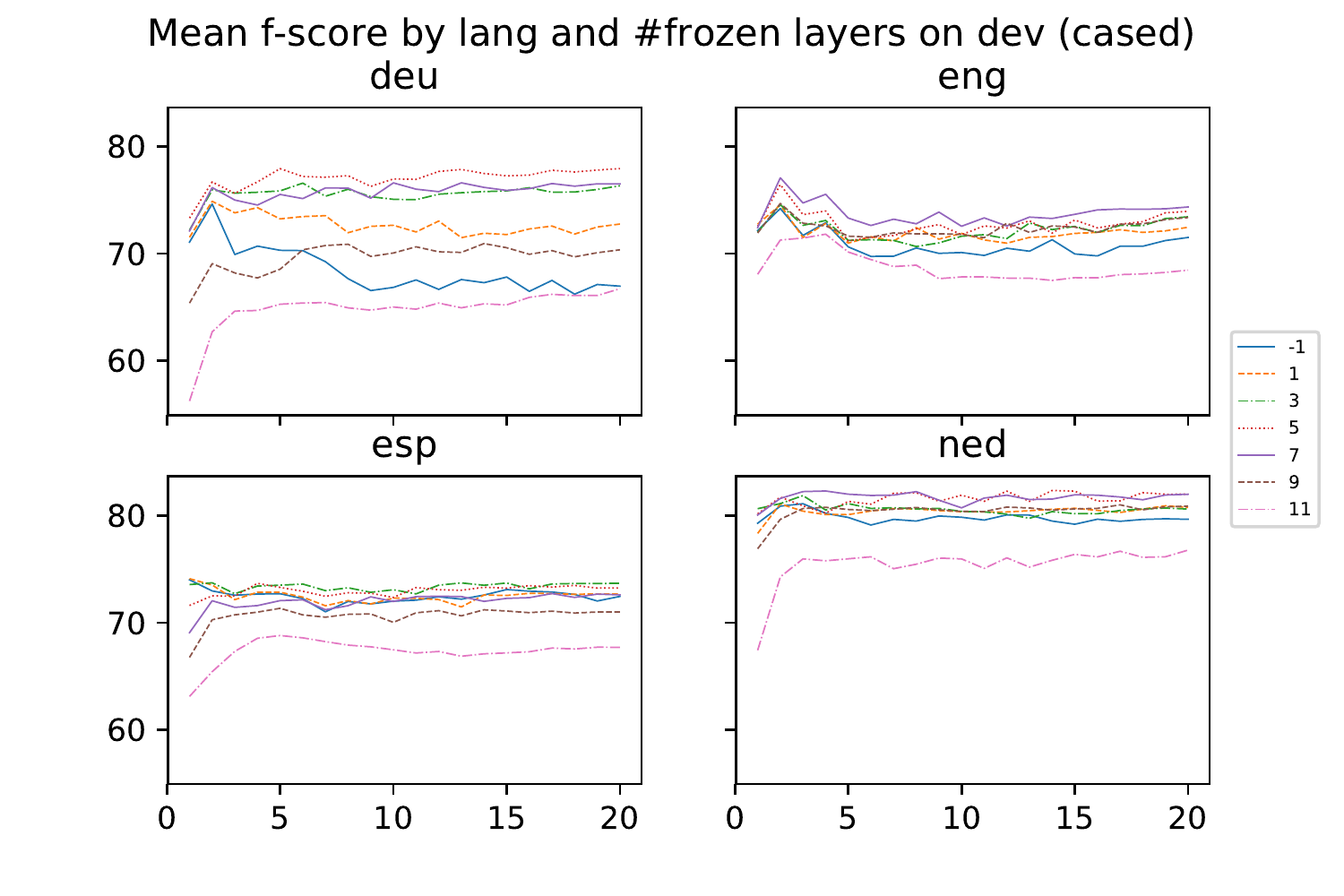}
\caption{Zero-shot BERT multilingual on CoNLL dev by language and epoch and number
  of frozen layers. Only odd frozen layers. Average of 5 runs}
\label{fig:cased_freeze_exps_n12_o}
\end{figure}


To overcome the pathological trajectory of the overall F-score as seen in Figure~\ref{fig:conll_fscore_epochs_by_lang_csd}, we propose freezing the lower layers of the BERT transformer architecture during learning so that they are not updated during training.\footnote{Dropout is also switched to evaluation
mode in the corresponding layers} We see this as a form of model regularization, and we conjecture that this is particularly useful in the zero-shot setting where the goal is to train an NER model on certain languages and to apply it to other languages that the model is not trained on. The underlying BERT representation will partially be the same as when it was trained on the multilingual Cloze task/consequent sentence prediction, while also partially being fit to the NER task.

We can see this conjecture bear out in Figure~\ref{fig:cased_freeze_exps_n12_o}.
The actual numerical results are presented in
Table \ref{tab:conll_frz_cased_dev} for CoNLL dev and table
\ref{tab:conll_frz_cased_tst2} for CoNLL test. 

During training, we freeze any of $-1, 0, 1,\ldots,12$ layers. Freezing -1 layer is identical to the baseline multilingual zero-shot learning setting in Section~\ref{sec:exp:zer} where nothing is frozen and everything is updated as in \cite{BERT18}. Freezing 0 layers means only the BERT embedding layer is frozen and none of the actual BERT layers are frozen. Freezing $1\ldots 12$ layers means that the embedding layer as well as the lowest $1\ldots 12$ layers are frozen during training and are not updated. The multilingual BERT embeddings use a 12 layer transformer. So when we freeze the lowest, say, 3 layers, only the subsequent layers from the fourth layer and on get updated as according the update regimen specified in \cite{BERT18}.

In fig.~\ref{fig:cased_freeze_exps_n12_o}, we show only the f-score plots for odd numbers of frozen layers ($-1,1,\ldots 11$) to reduce clutter. Focusing on the solid blue lines which correspond to the base zero-shot setting, the pathological behavior is most clearly visible in the German (deu) setting. However, when $1,3,\ldots 11$ layers are frozen, the plots of the f-scores stabilize and no longer decrease as learning progresses. The most stable trajectory occurs when 11 layers are frozen but underperform relative to when fewer layers are frozen.\footnote{though not shown in these graphs, freezing all 12 layers creates the most stable learning graph but with the lowest performance}

Tables~\ref{tab:conll_frz_cased_dev} and \ref{tab:conll_frz_cased_tst2} select and show some of the more interesting results when layers are frozen rather than showing results for all possible $-1,0,\ldots,12$ frozen layers. For both the dev and the test, we show that the best results are obtained when some layers are frozen with German, English and Dutch. Spanish is anomalous in this sense where no amount of freezing can outperform no freezing (i.e. -1). Furthermore, the scores for Spanish test are higher than for Spanish dev, which is highly unusual in that validation development scores are generally higher than held out test scores for most models. On dev, German performs best when 5 layers are frozen, English 7, and Dutch 5. On test, German performs best when 3 layers are frozen, English 7 and Dutch 3. \#frz 11 and 12 are presented in the tables to show that having too few layers to tune is detrimental in the zero-shot setting. Freezing 12 layers is analogous to the "feature based approach" in \cite{BERT18} where only the output layer is tuned for the CoNLL English task. There, the authors were able to achieve 91 f-score on the dev set without any fine tuning. Nonetheless, achieving 63.43 f-score for English (table~\ref{tab:conll_frz_cased_dev} when only the output layer is tuned on languages that are NOT English is still remarkable. This strongly supports our assumption that multilingual BERT is well-aligned and seems to generate a common representation for the 100+ language in a shared space.




%% file: 07-exps-mult.tex
\section{Experiments: Multi-task Learning}
\label{sec:exp:mult}

\input{tables/conll_exps_cased_tst2.tex}

Another possible solution to overcome the decreasing plot of the overall F-score as seen in Figure~\ref{fig:conll_fscore_epochs_by_lang_csd} is to pose zero-shot learning as a multitask learning problem.
Under the zero-shot assumption, we do not have any labeled data in the target language(s). However, it is reasonable to assume that we have raw unlabeled data in the target language and that we also know what languages the data is coming from. This allows us to utilize (1) the raw tokens in the target language and (2) the name of the target language.

Thus, we propose three additional tasks on top of the NER problem.
Let $(x_i^l, y_i^l), i \in \{1..n^l\}$ be the NER training data for language $l \in \{1..L\}$ for the $L$ languages where we have training data. Let $x_i^u, i \in \{1..n^u\}$ be the unlabeled raw sentences for language $u \in \{1..U\}$ for the $U$ languages where we do not have labeled data.

The training objective for the baseline zero-shot learning problem can be stated as follows:
\[
L(D_L) = \frac{1}{N} \sum_l^L \sum_i^{n^l} \ell(h^{\theta}(x_i^l), y_i^l)
\]
\noindent where $L$ is the total loss to minimize, $D_L$ is the labeled training data, $N$ is the total number of training instances, $\ell$ is the standard cross entropy loss and $h^{\theta}$ is the softmax output of the BERT encoding and a feed-forward layer and its associated parameters $\theta$.




\subsection{Language ID (LI)}

In the language ID (LI) multitask learning setting,
we add a secondary objective to identify the language of the unlabeled data. The
target language data is only used in the language identification
task. 
We use the dedicated initial classification token in BERT
(\texttt{'[CLS]'}) as the locus of the language ID layer. We use the
cross-entropy loss for the language ID. We sum this loss with the
baseline sequence prediction loss.
\[
L_{\mbox{LI}}(D_U) = \frac{1}{N_U} \sum_u^U \sum_i^{n^u} \ell(h_{\mbox{li}}^{\theta}(x_i^u), u)
\]
where $D_U$ is the set of unlabeled data, $N_U$ is the total number of unlabeled instances, $h_{\mbox{li}}^{\theta}$ is the output of the BERT embedding and a feed-forward layer for sentence classification along with the associated parameters $\theta$. Thus the loss function to minimize becomes
\[
L_{\mbox{LI}}(D_U) + L(D_L)
\]

\subsection{Cloze task (CL)}

A second variant is the Cloze task (CL) multitask learning objective, where
we add a secondary objective of language modeling to predict masked input wordpieces in the sentences of the target language as is used in the original BERT training objective~\cite{BERT18}.
The target language data has a subset of its input
wordpieces arbitrarily masked and the goal is to predict these masked wordpieces. We use the cross-entropy loss for the Cloze task. We sum this
loss with the baseline sequence prediction loss.
\[
L_{\mbox{CL}}(D_U) = \frac{1}{N_U} \sum_u^U \sum_i^{n^u} \ell(h_{\mbox{cl}}^{\theta}(m_p(x_i^u)), \hat{m}(x_i^u))
\]
where $h_{\mbox{cl}}^{\theta}$ is the output of the BERT embedding and a feed-forward layer for wordpiece prediction along with the associated parameters $\theta$. $m_p$ is a randomized masking function with a masking probability of $p$, and $\hat{m}$ is the associated output selection function which selects the original wordpieces which had been masked to the loss function. The loss function to minimize becomes
\[
L_{\mbox{CL}}(D_U) + L(D_L)
\]

\subsection{Predictive Cloze task (PC)}

The final variant is what we call the predictive Cloze task (PC) where we add a secondary objective to do a non-arbitrary Cloze task on the
target language data. 
Note that the NER model $h^{\theta}$ with the original set of parameters can be switched to decode input tokens at any moment during training. 
During training, we use the output of the
trained sequence labeler thus far to decode the target language
data. We mask the input wordpieces corresponding to the spans of predicted
entities and the goal is to predict these masked tokens. We use the
cross-entropy loss for the predictive Cloze task. We sum this loss
with the baseline sequence prediction loss.
\begin{flalign}
&L_{\mbox{PC}}(D_U)= \nonumber \\ 
&\frac{1}{N_U} \sum_u^U \sum_i^{n^u} \ell(h_{\mbox{pc}}^{\theta}(m(x_i^u,h^{\theta}(x_i^u)), \hat{m}(x_i^u,h^{\theta}(x_i^u))) \nonumber
\end{flalign}
where $D_U$ is the set of unlabeled data, $N_U$ is the total number of unlabeled instances, $h_{\mbox{pc}}^{\theta}$ is the output of the BERT embedding and a feed-forward layer for wordpiece prediction along with the associated parameters $\theta$. $m$ is a masking function which takes as input both the unlabeled text $x_i^u$ and the output of the NER decoder $h^{\theta}(x_i^u)$ to generate a masked input sequence over $x_i^u$, and $\hat{m}$ is the associated output selection function which selects the original wordpieces which had been masked by $m$ to the loss function.

\subsection{Combinations of secondary tasks}
In addition to the tasks laid out above, we consider settings where we combine the language ID (LI) task with either the Cloze (CL+LI) task or the predictive Cloze (PC+LI) task. By increasing the number of tasks for the challenge and thus increasing indirect regularization, we examine whether these further improve performance over the zero-shot setting.

\subsection{Experiment details}

In actual experiments, minibatch updates were used rather than batch updates as implied in the above formulations.
Due to the much heavier memory requirements brought on by the Cloze language modeling tasks, we reduce the maximum sequence length for the input sequences to 64. However, rather than truncating sentences and discarding the remainder, we segment long input sequences into smaller, overlapping sequences (with overlap size of 8) such that no training data is discarded. We use a masking probability $p=0.15$ as in the original BERT paper for the CL task.

All experiment results are averages over 5 runs with 5 different seeds. In terms of model selection for decoding the test data, the model with the best overall f-score on the respective development data was selected for each target language setting and seed.

\subsection{Experiment results}
We present results on the CoNLL test data for the above secondary tasks in table~\ref{tab:conll_exps_cased_tst2}. Here, the -1 value in the \#frz columns denote that no layers are frozen during training and are equivalent to the standard setting. At least for Dutch and English, we see that CL, CL+LI, LI, PC help as secondary tasks and improve over the ML(-1) baseline. However, there are no secondary tasks that help for Spanish. And only the LI task helps German. 

Next, we examine when 8 layers are frozen and the picture is mixed. For English, it helps with ML, LI, PC but doesn't seem to cause harm in the other cases. For Spanish, freezing 8 layers always hurts. For German and Dutch, there is no straightforward conclusion.

\input{tables/conll_csd_sl_tst.tex}
\subsubsection{Single-language zero-shot}
We also examine a combination of training a single language, namely CoNLL English and then assuming that there are no labeled data for Dutch, German and Spanish in Table~\ref{tab:conll_csd_sl_tst}.\footnote{We were unable to obtain results for certain settings where no BERT layers were frozen, i.e. the \#frz$=-1$ setting due to computational constraints.}

The results are presented analogously to Table~\ref{tab:conll_exps_cased_tst2}. The ML settings corresponds to the standard zero-shot setting but where the labeled training data comes from just one language. Here, the best scores occur when 3 layers are frozen.

Because we do not have results for when \#frz$=-1$ in the secondary tasks except for LI, it is difficult to say how they compare. However, it is possible to find settings where the -1 \#frz layer ML baseline is beaten, for example the 6 \#frz layer PC setting for Dutch, or the 8 \#frz layer LI setting for Spanish, or the 6 \#frz layer CL+LI setting for German.








%% file: tables/conll_exps_cased_tst2.tex
\begin{table}
\begin{center}
\begin{tabular}{llrrrr}
\toprule
 model & \#frz &   German &   English &  Spanish & Dutch \\
\midrule
ML & -1  & 70.34 & 72.57 & 77.17 & 79.76 \\ 
  & 8  & 69.94 & 74.29 & 75.19 & 80.60 \\      
CL & -1  & 67.04 & 73.30 & 75.29 & 81.76 \\
      &  8  & 67.37 & 73.03 & 73.11 & 81.23 \\
CL+LI & -1  & 68.94 & 74.28 & 73.68 & 80.78 \\
      &  8  & 66.16 & 74.42 & 73.85 & 79.97 \\
LI & -1  & 71.28 & 73.87 & 75.40 & 80.06 \\
      &  8  & 72.25 & 74.91 & 72.59 & 80.40 \\
PC & -1  & 67.91 & 73.72 & 73.68 & 80.28 \\
      &  8  & 67.29 & 75.68 & 73.37 & 79.95 \\
PC+LI & -1  & 66.31 & 74.79 & 74.12 & 79.53 \\
      &  8  & 65.47 & 74.79 & 73.59 & 80.35 \\
\bottomrule
\end{tabular}
\caption{CoNLL test results using ML, LI, CL, CL+LI and different
  frozen layers. It is possible that better numbers can be achieved with fewer than 8 frozen layers. The best results seem to occur around 3 frozen layers given table~\ref{tab:conll_frz_cased_tst2} but 8 frozen layers were chosen based on earlier, undiscussed experiments using uncased multilingual BERT with English.}
\label{tab:conll_exps_cased_tst2}
\end{center}
\end{table}

%% file: tables/conll_csd_sl_tst.tex
\begin{table}
\begin{center}
\begin{tabular}{llrrr}
\toprule
 Model     &  \#frz &   German &   Spanish &   Dutch \\
\midrule
ML & -1  & 69.42 & 73.62 & 78.61 \\
      &  3  & 71.42 & 75.67 & 80.38 \\ 
      &  6  & 70.16 & 73.74 & 78.20 \\
      &  8  & 65.45 & 73.14 & 77.05 \\
      &  12 & 34.68 & 53.72 & 52.56 \\
CL &  6  & 67.71 & 72.08 & 76.92 \\
      &  8  & 65.88 & 72.06 & 77.06 \\
CL+LI &  6  & 70.41 & 73.03 & 77.07 \\
      &  8  & 61.47 & 71.80 & 74.87 \\
LI & -1  & 69.83 & 72.26 & 74.47 \\
      &  6  & 65.65 & 72.91 & 76.37 \\
      &  8  & 68.90 & 73.92 & 77.92 \\
PC &  6  & 68.76 & 73.34 & 79.28 \\
      &  8  & 67.02 & 73.27 & 76.82 \\
PC+LI &  6  & 60.07 & 72.16 & 76.86 \\
      &  8  & 64.16 & 73.22 & 76.06 \\
\bottomrule
\end{tabular}
\caption{Single language zero shot with English CONLL on test using cased multilingual}
\label{tab:conll_csd_sl_tst}
\end{center}
\end{table}

%% file: 08-conc.tex
\section{Conclusion}

In this paper we present a simple and effective recipe of building multilingual NER systems with BERT. By utilizing a multilingual BERT framework, we were able to not only train a system that can perform inference on English, German, Spanish, and Dutch languages, but it performs better than the same model trained only on one language at a time, and also is able to perform 0-shot inference. The resulting model yields SotA results on CoNLL Spanish and Dutch, and on OntoNotes Chinese and Arabic datasets. In addition, the English trained model yields SotA results for 0-shot languages for Spanish, Dutch, and German NER, improving it by a range of 2.4F to 17.8F. Finally, the runtime signature (memory/CPU/GPU) of the model is the same as the models built on single languages, significantly simplifying its lifecycle maintenance.

We have shown that pretrained multilingual BERT embeddings can be used effectively in an unsupervised manner while harnessing annotated data outside the target language. Furthermore, it is easy to extend this approach through either partial gradient updates or the use of Cloze tasks and other unsupervised joint tasks to gain further improvements.







%% file: main.bbl
\begin{thebibliography}{}

\bibitem[\protect\citeauthoryear{Akbik, Blythe, and
  Vollgraf}{2018}]{akbik-etal-2018-contextual}
Akbik, A.; Blythe, D.; and Vollgraf, R.
\newblock 2018.
\newblock Contextual string embeddings for sequence labeling.
\newblock In {\em Proceedings of the 27th International Conference on
  Computational Linguistics},  1638--1649.
\newblock Santa Fe, New Mexico, USA: Association for Computational Linguistics.

\bibitem[\protect\citeauthoryear{Baevski \bgroup et al\mbox.\egroup
  }{2019}]{Baevski19-cloze5}
Baevski, A.; Edunov, S.; Liu, Y.; Zettlemoyer, L.; and Auli, M.
\newblock 2019.
\newblock Cloze-driven pretraining of self-attention networks.
\newblock {\em CoRR} abs/1903.07785.

\bibitem[\protect\citeauthoryear{Caruana}{1997}]{Caruana1997MultitaskL}
Caruana, R.
\newblock 1997.
\newblock Multitask learning.
\newblock {\em Machine Learning} 28:41--75.

\bibitem[\protect\citeauthoryear{Clark \bgroup et al\mbox.\egroup
  }{2018}]{clark-etal-2018-semi}
Clark, K.; Luong, M.-T.; Manning, C.~D.; and Le, Q.
\newblock 2018.
\newblock Semi-supervised sequence modeling with cross-view training.
\newblock In {\em Proceedings of the 2018 Conference on Empirical Methods in
  Natural Language Processing},  1914--1925.
\newblock Brussels, Belgium: Association for Computational Linguistics.

\bibitem[\protect\citeauthoryear{Collobert and Weston}{2008}]{collobert08}
Collobert, R., and Weston, J.
\newblock 2008.
\newblock A unified architecture for natural language processing: deep neural
  networks with multitask learning.
\newblock In {\em Proceedings of International Conference on Machine Learning}.

\bibitem[\protect\citeauthoryear{Devlin \bgroup et al\mbox.\egroup
  }{2018}]{BERT18}
Devlin, J.; Chang, M.; Lee, K.; and Toutanova, K.
\newblock 2018.
\newblock {BERT:} pre-training of deep bidirectional transformers for language
  understanding.
\newblock {\em CoRR} abs/1810.04805.

\bibitem[\protect\citeauthoryear{Duong \bgroup et al\mbox.\egroup
  }{2015}]{Duong2015LowRD}
Duong, L.; Cohn, T.; Bird, S.; and Cook, P.
\newblock 2015.
\newblock Low resource dependency parsing: Cross-lingual parameter sharing in a
  neural network parser.
\newblock In {\em ACL}.

\bibitem[\protect\citeauthoryear{Ghaddar and
  Langlais}{2018}]{ghaddar-langlais-2018-robust}
Ghaddar, A., and Langlais, P.
\newblock 2018.
\newblock Robust lexical features for improved neural network named-entity
  recognition.
\newblock In {\em Proceedings of the 27th International Conference on
  Computational Linguistics},  1896--1907.
\newblock Santa Fe, New Mexico, USA: Association for Computational Linguistics.

\bibitem[\protect\citeauthoryear{Github}{2019}]{akbik19:flair}
Github, Z.~R.
\newblock 2019.
\newblock very simple framework for state-of-the-art natural language
  processing (nlp).
\newblock https://github.com/zalandoresearch/flair.

\bibitem[\protect\citeauthoryear{{Google Research
  Github}}{2018}]{google-research18:_tensor_bert}
{Google Research Github}.
\newblock 2018.
\newblock Tensorflow code and pre-trained models for bert.
\newblock https://github.com/google-research/bert.

\bibitem[\protect\citeauthoryear{{H}ugging{Face}
  github}{2019}]{huggingface-github19}
{H}ugging{Face} github.
\newblock 2019.
\newblock The {B}ig-\&-{E}xtending-{R}epository-of-{T}ransformers: {P}retrained
  {P}y{T}orch models for {G}oogle's {BERT}, {OpenAI GPT \& GPT-2, Google/CMU
  Transformer-XL}.
\newblock https://github.com/huggingface/pytorch-pretrained-BERT.

\bibitem[\protect\citeauthoryear{Lample \bgroup et al\mbox.\egroup
  }{2016}]{LampleBSKD16}
Lample, G.; Ballesteros, M.; Subramanian, S.; Kawakami, K.; and Dyer, C.
\newblock 2016.
\newblock Neural architectures for named entity recognition.
\newblock {\em CoRR} abs/1603.01360.

\bibitem[\protect\citeauthoryear{Ni, Dinu, and
  Florian}{2017}]{ni-etal-2017-weakly}
Ni, J.; Dinu, G.; and Florian, R.
\newblock 2017.
\newblock Weakly supervised cross-lingual named entity recognition via
  effective annotation and representation projection.
\newblock In {\em Proceedings of the 55th Annual Meeting of the Association for
  Computational Linguistics (Volume 1: Long Papers)},  1470--1480.
\newblock Vancouver, Canada: Association for Computational Linguistics.

\bibitem[\protect\citeauthoryear{Peters \bgroup et al\mbox.\egroup
  }{2018}]{ELMo18}
Peters, M.~E.; Neumann, M.; Iyyer, M.; Gardner, M.; Clark, C.; Lee, K.; and
  Zettlemoyer, L.
\newblock 2018.
\newblock Deep contextualized word representations.
\newblock {\em CoRR} abs/1802.05365.

\bibitem[\protect\citeauthoryear{Pires, Schlinger, and
  Garrette}{2019}]{pires-etal-2019-multilingual}
Pires, T.; Schlinger, E.; and Garrette, D.
\newblock 2019.
\newblock How multilingual is multilingual {BERT}?
\newblock In {\em Proceedings of the 57th Annual Meeting of the Association for
  Computational Linguistics},  4996--5001.
\newblock Florence, Italy: Association for Computational Linguistics.

\bibitem[\protect\citeauthoryear{Pradhan \bgroup et al\mbox.\egroup
  }{2013}]{pradhan-etal-2013-towards}
Pradhan, S.; Moschitti, A.; Xue, N.; Ng, H.~T.; Bj{\"o}rkelund, A.; Uryupina,
  O.; Zhang, Y.; and Zhong, Z.
\newblock 2013.
\newblock Towards robust linguistic analysis using {O}nto{N}otes.
\newblock In {\em Proceedings of the Seventeenth Conference on Computational
  Natural Language Learning},  143--152.
\newblock Sofia, Bulgaria: Association for Computational Linguistics.

\bibitem[\protect\citeauthoryear{sil}{2015}]{sil2015-tac}
2015.
\newblock {\em Proceedings of the 2015 Text Analysis Conference, {TAC} 2015,
  Gaithersburg, Maryland, USA, November 16-17, 2015, 2015}, {NIST}.

\bibitem[\protect\citeauthoryear{{Tjong~Kim~Sang} and
  Veenstra}{1999}]{sang99representing}
{Tjong~Kim~Sang}, E.~F., and Veenstra, J.
\newblock 1999.
\newblock Representing text chunks.
\newblock In {\em Proceedings of EACL'99}.

\bibitem[\protect\citeauthoryear{Tjong
  Kim~Sang}{2002}]{TjongKimSang:2002:ICS:1118853.1118877}
Tjong Kim~Sang, E.~F.
\newblock 2002.
\newblock Introduction to the conll-2002 shared task: Language-independent
  named entity recognition.
\newblock In {\em Proceedings of the 6th Conference on Natural Language
  Learning - Volume 20}, COLING-02,  1--4.
\newblock Stroudsburg, PA, USA: Association for Computational Linguistics.

\bibitem[\protect\citeauthoryear{Weischedel \bgroup et al\mbox.\egroup
  }{2011}]{ontonotes5}
Weischedel, R.; Hovy, E.; Marcus, M.; Palmer, M.; Belvin, R.; Pradhan, S.;
  Ramshaw, L.; and Xue, N.
\newblock 2011.
\newblock {\em OntoNotes: A Large Training Corpus for Enhanced Processing}.

\bibitem[\protect\citeauthoryear{Xie \bgroup et al\mbox.\egroup
  }{2018}]{Xie2018}
Xie, J.; Yang, Z.; Neubig, G.; Smith, N.~A.; and Carbonell, J.
\newblock 2018.
\newblock Neural cross-lingual named entity recognition with minimal resources.
\newblock In {\em Proceedings of the 2018 Conference on Empirical Methods in
  Natural Language Processing},  369--379.
\newblock Brussels, Belgium: Association for Computational Linguistics.

\end{thebibliography}
